\documentclass[10pt,twocolumn,letterpaper]{article}

\usepackage{wacv}
\usepackage{times}
\usepackage{epsfig}
\usepackage{graphicx}
\usepackage{amsmath}
\usepackage{amssymb}

\usepackage[pagebackref=true,breaklinks=true,letterpaper=true,colorlinks,bookmarks=false]{hyperref}

\wacvfinalcopy 

\def\wacvPaperID{118} 

\ifwacvfinal\fi
\begin{document}

\title{Image Difficulty Curriculum for Generative Adversarial Networks (CuGAN)\vspace{-0.4cm}}

\author{Petru Soviany$^{1}$, Claudiu Ardei$^1$, Radu Tudor Ionescu$^1$, Marius Leordeanu$^{2,3}$\\
$^1$University of Bucharest\\
$^2$Institute of Mathematics of the Romanian Academy, $^3$University ``Politehnica" of Bucharest\\
\vspace{-0.8cm}
}

\maketitle

\begin{abstract}
\vspace{-0.3cm}
Despite the significant advances in recent years, Generative Adversarial Networks (GANs) are still notoriously hard to train. In this paper, we propose three novel curriculum learning strategies for training GANs. All strategies are first based on ranking the training images by their difficulty scores, which are estimated by a state-of-the-art image difficulty predictor. Our first strategy is to divide images into gradually more difficult batches. Our second strategy introduces a novel curriculum loss function for the discriminator that takes into account the difficulty scores of the real images. Our third strategy is based on sampling from an evolving distribution, which favors the easier images during the initial training stages and gradually converges to a uniform distribution, in which samples are equally likely, regardless of difficulty.
We compare our curriculum learning strategies with the classic training procedure on two tasks: image generation and image translation. Our experiments indicate that all strategies provide faster convergence and superior results. For example, our best curriculum learning strategy applied on spectrally normalized GANs (SNGANs) fooled human annotators in thinking that generated CIFAR-like images are real in $25.0\%$ of the presented cases, while the SNGANs trained using the classic procedure fooled the annotators in only $18.4\%$ cases. Similarly, in image translation, the human annotators preferred the images produced by the Cycle-consistent GAN (CycleGAN) trained using curriculum learning in $40.5\%$ cases and those produced by CycleGAN based on classic training in only $19.8\%$ cases, $39.7\%$ cases being labeled as ties.
\vspace{-0.4cm}
\end{abstract}

\setlength{\abovedisplayskip}{4pt}
\setlength{\belowdisplayskip}{4pt}

\section{Introduction}
\vspace{-0.1cm}
Generative Adversarial Networks (GANs)~\cite{Goodfellow-NIPS-2014} represent a hot topic in computer vision, drawing the attention of many researchers who developed several improvements of the standard architecture~\cite{Arjovsky-arXiv-2017,Choi-CVPR-2018,Gulrajani-NIPS-2017,Isola-CVPR-2017,Lin-NIPS-2018,Mao-ICCV-2017,Mirza-arXiv-2014,Odena-ICML-2017,Radford-ICLR-2016,Reed-ICML-2016,Tolstikhin-NIPS-2017,Wang-ECCV-2016}. Yet, this kind of neural models are still very hard to train~\cite{Mescheder-NIPS-2017}. In this paper, we study the hypothesis of improving the training process of GANs in terms of both accuracy and time, by employing curriculum learning~\cite{Bengio-ICML-2009}. \emph{Curriculum learning} is the process of training machine learning models by presenting the training examples in a meaningful order which gradually illustrates more complex concepts. Although many curriculum learning approaches~\cite{Gong-TIP-2016,Graves-ICML-2017,Gui-FG-2017,Ionescu-CVPR-2016,Jiang-ICML-2018,Li-BMVC-2017,Morerio-ICCV-2017,Wang-ICPR-2018,Wang-ECCV-2018,Zhang-IJCV-2019} have been proposed for training deep neural networks, to our knowledge, there are only a few studies that apply curriculum learning to GANs~\cite{Doan-AAAI-2019,Ghasedi-CVPR-2019}.

\begin{figure}[!t]
\begin{center}
\includegraphics[width=0.63\linewidth]{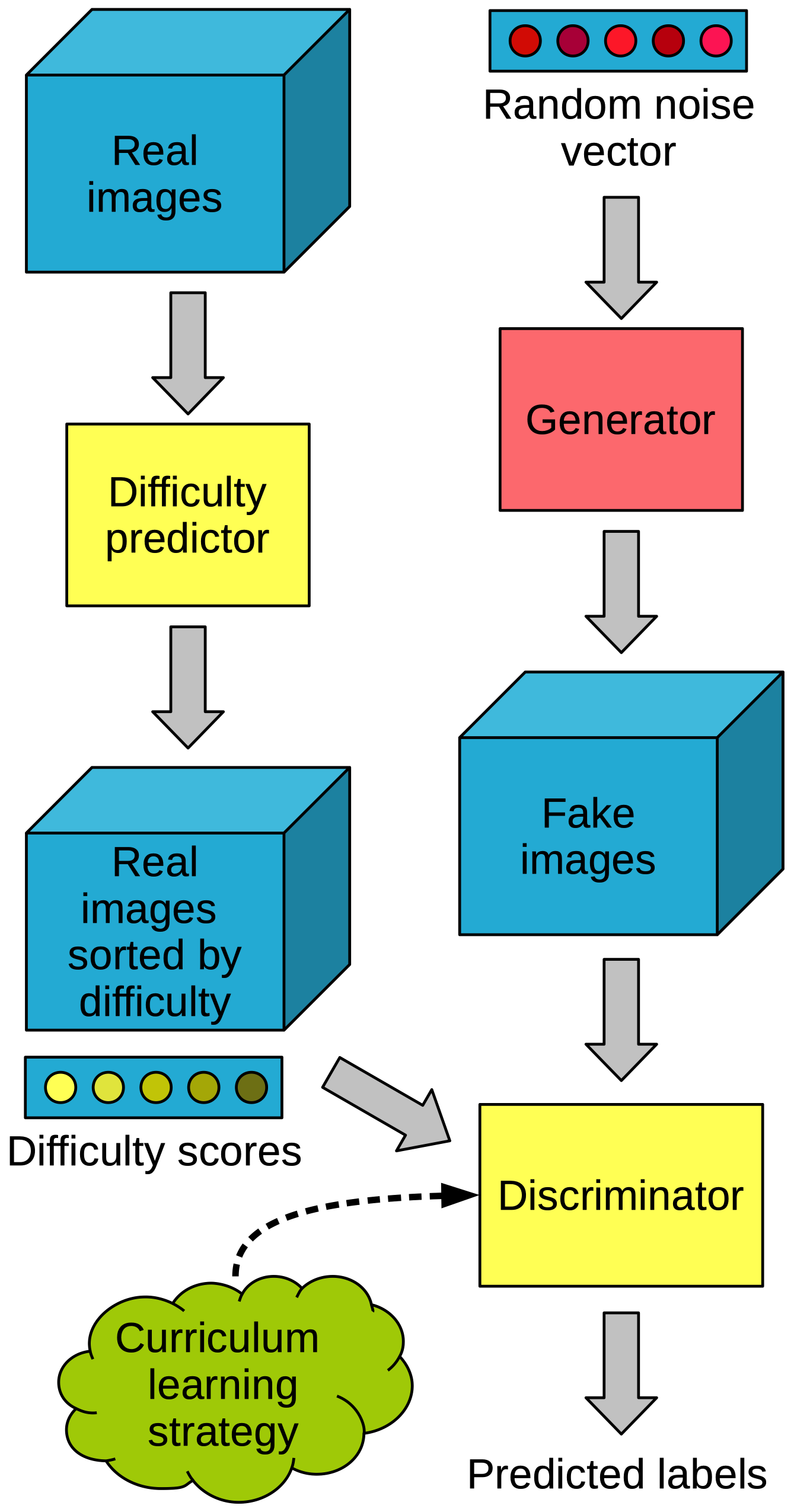}
\end{center}
\vspace*{-0.4cm}
\caption{Our GAN training pipeline based on curriculum learning. The real training images are passed to an image difficulty predictor which provides a difficulty score for each image. A curriculum learning strategy that takes into account the difficulty scores is employed to train the discriminator. 
Best viewed in color.}
\label{fig_pipeline}
\vspace*{-0.5cm}
\end{figure}

In this paper, we propose three novel curriculum learning strategies that provide faster convergence during GANs training, as well as improved results. Our curriculum learning strategies are general enough to be applied to any GAN architecture, as shown in Figure~\ref{fig_pipeline}. They rely on a state-of-the-art image difficulty predictor~\cite{Ionescu-CVPR-2016}, which scores the (real) training images with respect to the difficulty of solving a visual search task. After receiving the image difficulty scores as input, we employ one of our curriculum learning strategies listed below:
\begin{itemize}
\vspace{-0.15cm}
\item Divide the training images into $m$ easy-to-hard batches and start training the GAN with the easy batch. The other batches are added into the training process, in increasing order of difficulty, after a certain number of iterations.
\vspace{-0.15cm}
\item Add another component to the discriminator loss function which makes the loss value proportional to the easiness (inverse difficulty) score of the images. The impact of this new component is gradually attenuated, until the easiness score has no more influence in the last training iterations.
\vspace{-0.15cm}
\item Change the discriminator loss function by including probabilities of sampling real images from a biased distribution that strongly favors easier images during the first training iterations. The probability distribution is continuously updated with each iteration, until it becomes uniform in the last training iterations.
\end{itemize}
\vspace{-0.15cm}

Our three curriculum learning strategies follow two important principles. First, we keep the easier images until the end of the training process, to prevent catastrophic forgetting~\cite{Kirkpatrick-PNAS-2017,McCloskey-PLM-1989}. Second, we want all training examples to receive equal importance \emph{in the end} (when training is finished), as we have no reason to favor the easy or the difficult images. However, during the initial stages of training, we emphasize easier images in order to achieve faster convergence and possibly a better local minimum.

We perform image generation experiments using the spectrally normalized GAN (SNGAN) model~\cite{Miyato-ICLR-2018}, and image translation experiments using the Cycle-consistent GAN (CycleGAN) model~\cite{Zhu-ICCV-2017}. The goal of our experiments is to compare the standard training process, in which examples are presented in a random order, with the training process based on curriculum. The image generation results on CIFAR-10~\cite{Krizhevsky-TR-2009} indicate that all the proposed curriculum learning strategies improve the Inception Score (IS)~\cite{Salimans-NIPS-2016} and the Fr\'echet Inception Distance (FID)~\cite{Heusel-NIPS-2017} over the state-of-the-art SNGAN model. Furthermore, we conducted several human annotations studies, to determine whether our generated or translated images are better than those produced by the baselines SNGAN and CycleGAN, respectively. Our best curriculum learning strategy fooled human annotators in thinking that generated CIFAR-like images are real in $25.0\%$ of the presented cases (on average), while the SNGAN fooled the annotators in only $18.4\%$ cases. This represents an absolute gain of $6.6\%$ over SNGAN. We obtain significant improvements in image translation as well. For example, in the horse2zebra~\cite{Zhu-ICCV-2017} experiment, the human annotators opted for our method in $52.5\%$ of the presented cases and for the baseline CycleGAN in only $11.9\%$ cases, $35.6\%$ cases being labeled as draws. This represents an absolute gain of $40.6\%$ over CycleGAN. We thus conclude that employing curriculum learning for training GANs is useful.

We organize the rest of this paper as follows. In Section~\ref{sec_Related_Work}, we present related works and how our approach is different.
In Section~\ref{sec_Method}, we describe our curriculum learning strategies for training GANs. We present the image generation and image translation experiments in Section~\ref{sec_Experiments}. We draw our conclusion and discuss future work in Section~\ref{sec_Conclusion}.

\vspace{-0.2cm}
\section{Related work}
\label{sec_Related_Work}

\noindent
{\bf Generative Adversarial Networks.}
Generative Adversarial Networks~\cite{Goodfellow-NIPS-2014} are composed of two neural networks, a generator and a discriminator, which are trained for generating new images, similar to those provided in a training set. Since 2014, many variations of GANs have been proposed in order to improve the quality of the generated samples~\cite{Arjovsky-arXiv-2017,Choi-CVPR-2018,Gulrajani-NIPS-2017,Isola-CVPR-2017,Lin-NIPS-2018,Mao-ICCV-2017,Mirza-arXiv-2014,Odena-ICML-2017,Radford-ICLR-2016,Reed-ICML-2016,Tolstikhin-NIPS-2017,Wang-ECCV-2016}. Mirza and Osindero \cite{Mirza-arXiv-2014} introduced a conditional version of GANs, termed CGAN, which is based on feeding label information to both the generator and the discriminator. As CGAN, the Auxiliary Classifier GAN (AC-GAN) \cite{Odena-ICML-2017} is a class conditional model, which in addition, leverages side information through an auxiliary decoder that is responsible for reconstructing class labels. Deep convolutional GANs (DCGANs) \cite{Radford-ICLR-2016} include a set of constraints to the architectural topology of the classic GAN, to improve training stability. Wasserstein GANs (WGANs) \cite{Arjovsky-arXiv-2017} use the Earth Mover distance instead of other popular metrics to provide easier training, while lowering the chances of entering mode collapse. Still, WGAN employs a weight clipping technique which can result in failure to converge and bad outputs. This problem is addressed in WGAN-GP (Wasserstein GAN with Gradient Penalty) \cite{Gulrajani-NIPS-2017}, where weight clipping is replaced by gradient penalty, providing better performance on different architectures. SNGAN~\cite{Miyato-ICLR-2018} introduces spectral normalization, another normalization technique used to stabilize the training of the discriminator. Compared to the other regularization methods, spectral normalization provides better results and lower computational costs. 


CycleGAN~\cite{Zhu-ICCV-2017} performs image translation without requiring paired images to learn the mapping. Instead, it learns the relevant features of two domains and how to translate between these domains. It uses cycle consistency, which encodes the idea that translating from one domain to another and back again should take you back to the same place. Choi et al.~\cite{Choi-CVPR-2018} introduced StarGAN, a conditional solution that has the advantage of providing good results when translating between more than two domains, using a single discriminator and generator network. 

Some studies \cite{Isola-CVPR-2017,Reed-ICML-2016,Wang-ECCV-2016} showed that providing additional information to a GAN model can result in performance improvements in a wide range of common generative tasks. Similar to these approaches, we use an external difficulty score, trying to constrain the order and the importance of the training samples, in order to imitate the easy-to-hard (curriculum) learning paradigm from humans. 


\noindent
{\bf Curriculum learning.}
Bengio et al.~\cite{Bengio-ICML-2009} studied easy-to-hard strategies to train machine learning models, showing that machines can also benefit from learning by gradually adding more difficult examples. They introduced a general formulation of the easy-to-hard training strategies known as \emph{curriculum learning}. In the past few years curriculum learning has been applied to semi-supervised image classification~\cite{Gong-TIP-2016}, language modelling~\cite{Graves-ICML-2017}, question answering~\cite{Graves-ICML-2017}, object detection~\cite{Soviany-CEFRL-2018,Soviany-SYNASC-2018,Wang-ICPR-2018,Zhang-IJCV-2019}, person re-identification~\cite{Wang-ECCV-2018}, weakly supervised object detection~\cite{Ionescu-CVPR-2016,Li-BMVC-2017}. Other works proposed refined techniques for improving neural network training components, e.g. dropout~\cite{Morerio-ICCV-2017}, or training frameworks, e.g. teacher-student~\cite{Jiang-ICML-2018}, using curriculum learning.
Ionescu et al.~\cite{Ionescu-CVPR-2016} considered an image difficulty predictor trained on image difficulty scores produced by human annotators. Similar to Ionescu et al.~\cite{Ionescu-CVPR-2016}, we use an image difficulty predictor, but with a completely different purpose, that of training GANs. In addition, we explore several curriculum learning strategies that enable end-to-end training by defining new curriculum loss functions.

\noindent
{\bf Curriculum GANs.}
To our knowledge, there are a just few works that propose curriculum learning approaches for training GANs~\cite{Doan-AAAI-2019,Ghasedi-CVPR-2019}. Doan et al.~\cite{Doan-AAAI-2019} introduced an adaptive curriculum learning strategy for training GANs, called acGAN, which uses multiple discriminators with different architectures of various depths. The authors proposed a reward function that uses an online multi-armed bandit algorithm. The reward function measures the progress made by the generator and uses it to update the weights of each discriminator, ensuring that the generator and the discriminators learn at the same pace. Different from the approach of Doan et al.~\cite{Doan-AAAI-2019}, we consider the difficulty of the training samples and propose strategies to train GANs gradually, from the easy images to the hard ones. While the approach of Doan et al.~\cite{Doan-AAAI-2019} uses multiple discriminators, increasing the training time, our approach does not require any additional training time. 

Ghasedi et al.~\cite{Ghasedi-CVPR-2019} proposed ClusterGAN, an easy-to-difficult approach for image clustering. ClusterGAN is an unsupervised model composed of three elements: a generator, a discriminator and a clustering network. The samples are introduced gradually in the training, from the easy ones to the hard ones. The values of the loss function are used as difficulty scores for the corresponding image samples. Their curriculum learning strategy leads to good results when training clustering networks with large depth. While Ghasedi et al.~\cite{Ghasedi-CVPR-2019} study the problem of clustering images, we apply curriculum learning in order to generate or translate images. Furthermore, we propose and study three curriculum learning strategies instead of a single one.

\vspace{-0.1cm}
\section{Method}
\label{sec_Method}
\vspace{-0.1cm}
\subsection{Preliminaries and notations}
\vspace{-0.1cm}

Generative Adversarial Networks~\cite{Goodfellow-NIPS-2014} are composed of two neural networks, the generator (G) and the discriminator (D), which are trained to compete against each other in an adversarial game. The generator learns to generate image samples from a Gaussian noise density $p_z$, such that the generated (fake) images (from the learned density $p_g$) are difficult to distinguish from real images for the discriminator. Meanwhile, the discriminator is trained to differentiate between real images from a density $p_r$ and fake images from the density $p_g$ learned by G. The two networks, G and D, compete in a minimax game with the objective function $V(G, D)$ defined as follows:
\begin{equation}\label{eq_gan}
V(G,D) = \mathbb{E}_{x \sim p_r}[l(D(x))] + \mathbb{E}_{z \sim p_z}[l(-D(G(z)))],
\end{equation}
where $x$ is a real image sampled from the true data density $p_r$, $z$ is the random noise vector sampled from the density $p_z$, and $l$ is a loss function, e.g. cross-entropy~\cite{Goodfellow-NIPS-2014} or Hinge loss~\cite{Miyato-ICLR-2018}. The goal of the generator $G$ is to minimize this error, while the goal of the discriminator $D$ is to maximize it. Hence, during training, we aim to optimize the objective function as follows:
\begin{equation}
\min_{G} \max_{D} V(G,D).
\end{equation}
The two networks are alternatively trained until the generator learns the probability density function of the training data $p_r$, i.e. until $p_g \approx p_r$.

\vspace{-0.1cm}
\subsection{Curriculum GANs based on image difficulty}
\label{sec_Method_sub}
\vspace{-0.1cm}

While machines are commonly trained by presenting examples in a random order, humans learn new concepts by organizing them in a meaningful order which gradually illustrates higher complexity. To this end, Bengio et al.~\cite{Bengio-ICML-2009} proposed curriculum learning for training machine learning models, specifically neural networks, which are influenced by the order in which the examples are presented during training. Since deep neural networks are models that essentially try to mimic the brain, it seems natural to also adopt curriculum learning from humans~\cite{Smith-TCS-2018}. 
We hypothesize that a curriculum learning strategy for training GANs can bring several benefits, e.g. faster convergence, improved stability and superior results. To demonstrate our hypothesis we explore three curriculum learning strategies that are generic enough to be applied to any GAN architecture. In order to learn in the increasing order of difficulty (from easy to hard), we first need to apply an image difficulty predictor on the training set of real images. This allows us to change the distribution of the real images $p_r$ in order to introduce curriculum when training the discriminator D. Since the generator G tries to learn a distribution $p_g$ that closely follows $p_r$, G is implicitly influenced by the curriculum learning strategy. Therefore, it is not necessary to apply the image difficulty predictor on the generated images, saving the additional training time. Moreover, the difficulty predictor needs to be applied only once on the real images, before starting to train the GANs. We next present the image difficulty predictor and our three curriculum learning strategies.

\begin{figure}[!t]
\begin{center}
\includegraphics[width=1.0\linewidth]{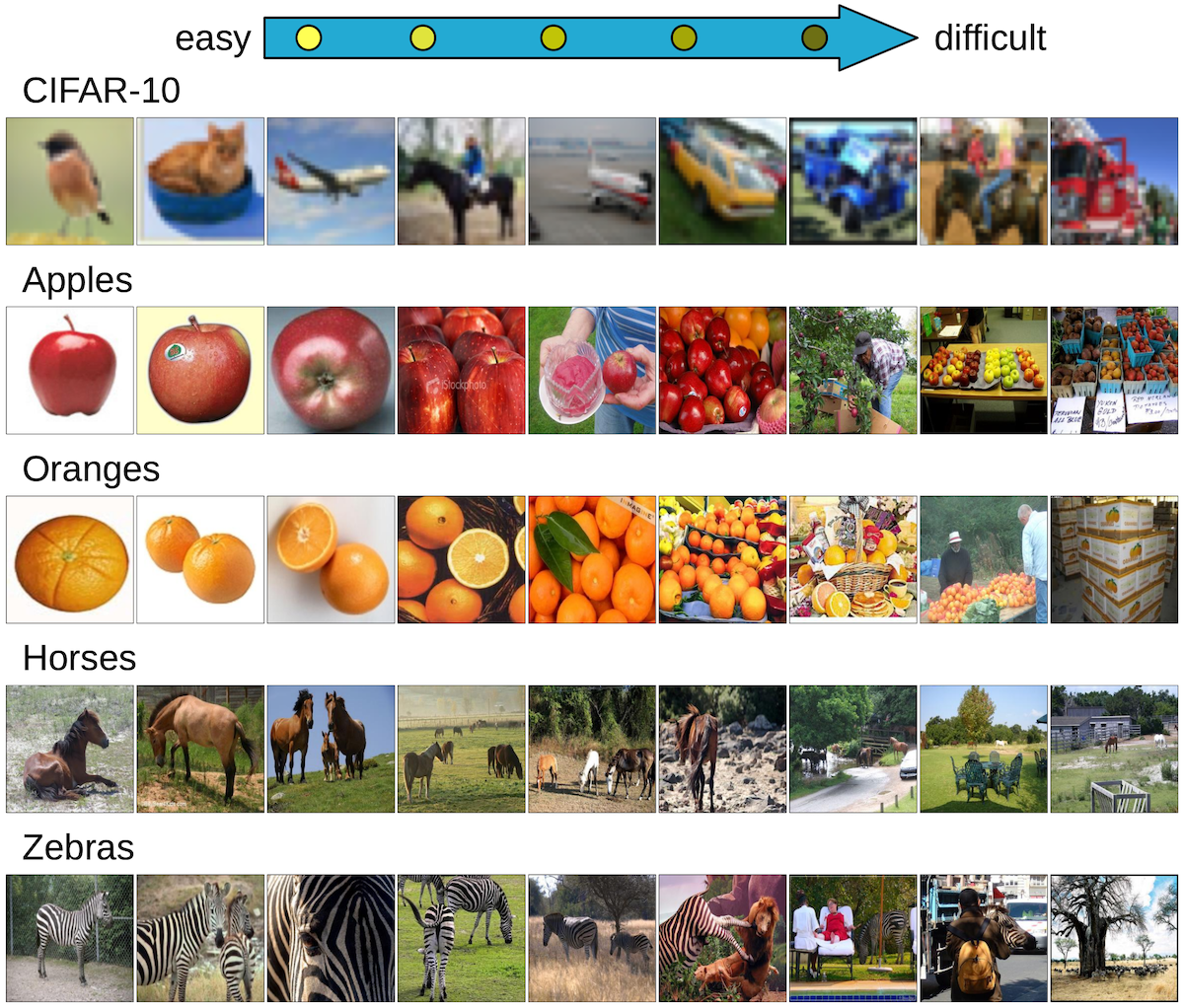}
\end{center}
\vspace*{-0.3cm}
\caption{From left to right, images in increasing order of difficulty selected from CIFAR-10~\cite{Krizhevsky-TR-2009}, apple2orange~\cite{Zhu-ICCV-2017} and horse2zebra~\cite{Zhu-ICCV-2017} data sets, respectively. Best viewed in color.}
\label{fig_easy_to_hard}
\vspace*{-0.4cm}
\end{figure}

\noindent
{\bf Image difficulty prediction.}
Ionescu et al.~\cite{Ionescu-CVPR-2016} defined  image difficulty as the human response time for solving a visual search task, collecting corresponding difficulty scores for the PASCAL VOC 2012 data set~\cite{Pascal-VOC-2012}. We follow the approach proposed in~\cite{Ionescu-CVPR-2016} to build a state-of-the-art image difficulty predictor. The model is based on concatenating deep features extracted from two Convolutional Neural Networks (CNN), VGG-f~\cite{Chatfield-BMVC-14} and VGG-verydeep-16~\cite{Simonyan-ICLR-14}, which are pre-trained on ImageNet~\cite{Russakovsky2015}. We remove the softmax layer of each CNN model and use the output of the penultimate fully-connected layer, resulting in a feature vector of $4096$ components. We divide each image into $1 \times 1$, $2 \times 2$ and $3 \times 3$ bins in order to obtain a pyramid representation, which leads to performance improvements~\cite{Ionescu-CVPR-2016}. We concatenate the feature vectors corresponding to each bin into a single vector corresponding to the entire image. We $L_2$-normalize the concatenated feature vectors before training a $\nu$-Support Vector Regression ($\nu$-SVR)~\cite{Chang-NC-2002} model to regress to the ground-truth difficulty scores provided for PASCAL VOC 2012~\cite{Pascal-VOC-2012}. We use the learned predictor $P$ as an image difficulty scoring function that provides difficulty scores on a continuous scale:
\begin{equation}\label{eq_difficulty_predictor}
s_i = \frac{P(x_i) -  \min_{x_j \in X}\{P(x_j)\}}{\max_{x_j \in X}\{P(x_j)\}} \cdot 2 - 1,
\end{equation}
where $s_i$ is the difficulty score for the image $x_i$ in a set of images $X = \{x_1, x_2, ...,x_n\}$, where $n=|X|$. Eq.~\eqref{eq_difficulty_predictor} maps the predicted difficulty scores for the set $X$ to the interval $[-1, 1]$. Our predictor attains a Kendall's $\tau$ correlation coefficient of $0.471$ on the same test set of~\cite{Ionescu-CVPR-2016}. In Figure~\ref{fig_easy_to_hard}, we present images in increasing order of difficulty from the data sets considered in our experiments from Section~\ref{sec_Experiments}.

\noindent
{\bf Learning using image difficulty batches.}
Our first curriculum learning strategy is based on dividing the real images into $m$ equally-sized batches indexed from $1$ to $m$, of increasing difficulty, such that images in each batch $i+1$ have higher difficulty scores than the images in the batch $i$, $\forall i \in \{1,2,...,m-1\}$. Thus, the first batch contains the easiest images and the last batch contains the hardest ones. After dividing the images into batches of increasing difficulty, we start training the GANs using only images from the first batch. After a fixed number of iterations, we include images from the second batch into the training. This process continues until all $m$ batches are included into the training. In this way, the generator learns a progressively complex density $p_g$. Since images in the former batches can be learned faster (due to their easiness), we consider a smaller number of iterations during the early training stages. The number of iterations increases as more difficult batches are added.

\begin{figure}[!t]
\begin{center}
\includegraphics[width=0.98\linewidth]{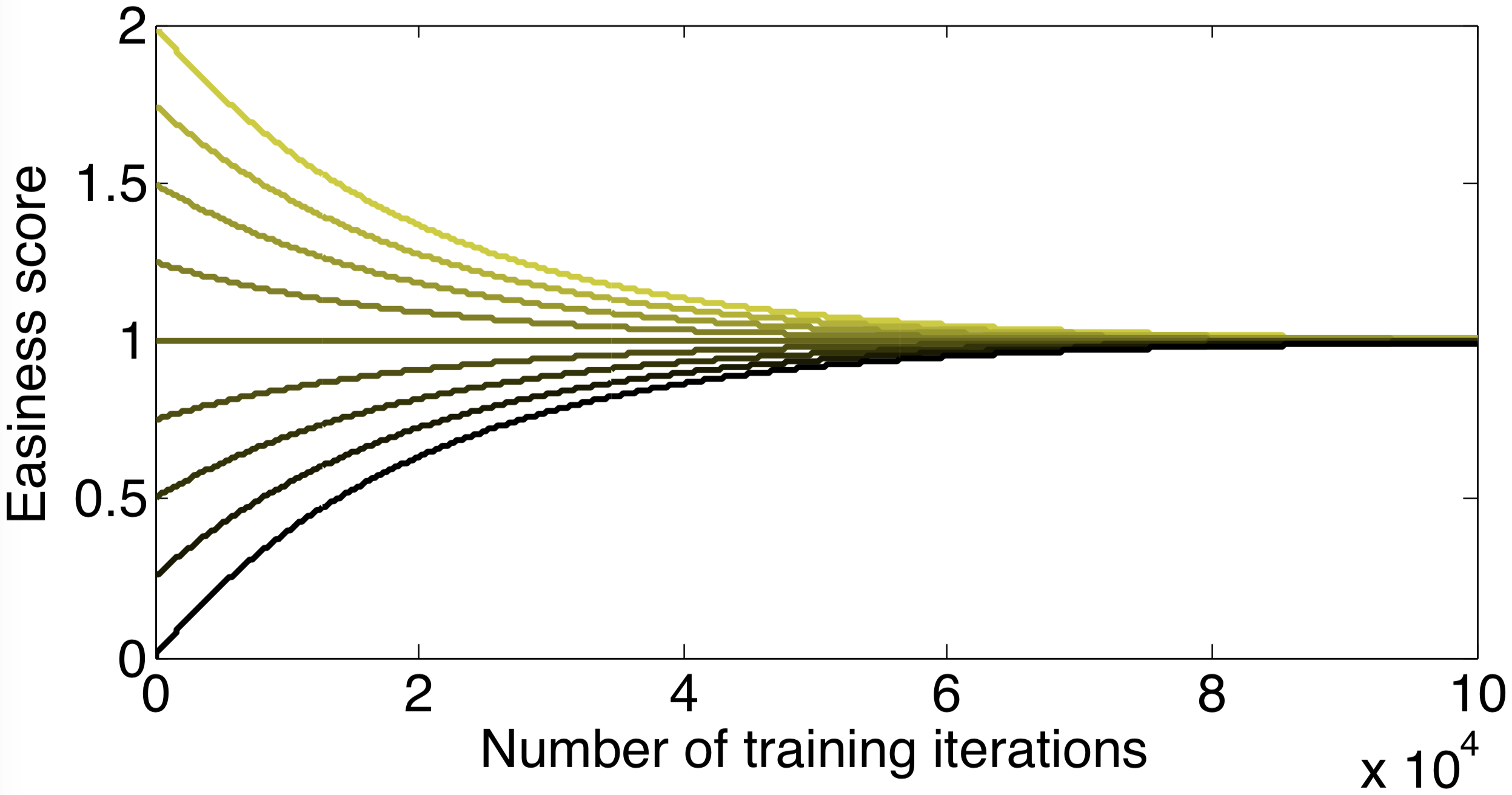}
\end{center}
\vspace*{-0.3cm}
\caption{Easiness scores between $0$ and $2$ converge to $1$ as the number of training iterations increases, by applying the scoring function defined in Eq.~\eqref{eq_scoring_function} with $k=1$ and $\gamma = 5 \cdot 10^{-5}$. Each curve represents the evolution of the weight for a given image, which starts with the weight equal to its easiness score $(1-s)$ at the first iteration and ends with a weight equal to 1, regardless of its initial easiness score. Best viewed in color.}
\label{fig_scoring_function}
\vspace*{-0.4cm}
\end{figure}

\noindent
{\bf Learning by weighting according to difficulty.}
Our second curriculum learning strategy is based on integrating the difficulty scores into the discriminator loss function, by weighting the real images according to their difficulty scores. In the first training iterations, we aim to provide higher weights to the easy images and lower weights to the difficult images. With each training iteration, the weights of both easy and difficult images gradually converge to a single value. The weights are computed using the following scoring function $w_P$:
\begin{equation}\label{eq_scoring_function}
w_P(x_i,t) = 1 - k \cdot s_i \cdot e^{-\gamma \cdot t},
\end{equation}
where $x_i$ is an image from the set of real images $X$, $s_i$ is the image difficulty score as in Eq.~\eqref{eq_difficulty_predictor}, $t$ is the current training iteration index, $\gamma$ is a parameter that controls how fast the scores converge to the value $1$ and $k$ is a parameter that controls the impact of the difficulty weights to the overall loss value. Figure~\ref{fig_scoring_function} illustrates the behavior of $w_P$ for various easiness scores in the interval $[0, 2]$. In the first iteration (when $t=0$), the easiness scores are equal to $1 - s$, for $k=1$. Note that in the last iterations all images have basically the same weight, regardless of their difficulty. By including the scoring function $w_P$ from Eq.~\eqref{eq_scoring_function} into the objective function $V$ defined in Eq.~\eqref{eq_gan}, we obtain a novel objective (loss) function $V^{(1)}$ based on curriculum learning, defined as follows:
\begin{equation}\label{eq_curriculum2}
\begin{split}
V^{(1)}(G,D,P) &= \mathbb{E}_{x \sim p_r}[l(D(x)) + w_P(x,t)]\\
&+ \mathbb{E}_{z \sim p_z}[l(-D(G(z)))].
\end{split}
\end{equation}
We note that when $t$ approaches infinity, the objective function $V^{(1)}$ converges asymptotically to $V + 1$, i.e.:
\begin{equation}
\lim_{t \rightarrow{\infty}}V^{(1)}(G,D,P) = V(G,D) + 1.
\end{equation}
This can be immediately demonstrated by considering that:
\begin{equation}\label{eq_scoring_convergence}
\lim_{t \rightarrow{\infty}} w_P(x,t) = \lim_{t \rightarrow{\infty}} 1 - k \cdot s \cdot e^{-\gamma \cdot t} = 1.
\end{equation}

\noindent
{\bf Learning by sampling according to difficulty.}
Our third curriculum learning strategy is based on changing the probability density function of the real images $p_r$, by multiplying it with another probability density function that is proportional to $w_P$ defined in Eq.~\eqref{eq_scoring_function}:
\begin{equation}
p_{r,w_P} = p_r \cdot p_{w_P} \propto w_P(x,t).
\end{equation}
By including the novel density $p_{r,w_P}$ into the objective function $V$ defined in Eq.~\eqref{eq_gan}, we effectively obtain a novel loss function $V^{(2)}$ based on curriculum learning, defined as follows:
\begin{equation}\label{eq_curriculum3}
\begin{split}
V^{(2)}(G,D,P) &= \mathbb{E}_{x \sim p_{r,w_P}}[l(D(x))]\\
&+ \mathbb{E}_{z \sim p_z}[l(-D(G(z)))].
\end{split}
\end{equation}

We use the weights $w_P(x_i,t)$ to define a distribution over the training images. We then sample training images from this distribution during training. We define a discrete random variable $R$ with possible values associated to indexes of images in the training set $X$, such that the probability $Prob(R=i)$ of sampling an index for real image $x_i$ from $X$ is equal to the weight $w_P(x_i,t)$ 
divided by the sum of all weights, making all probabilities sum up to $1$:
\begin{equation}\label{eq_prob}
Prob(R=i) = \frac{w_P(x_i,t)}{\sum_{x_j \in X} w_P(x_j,t)}, \forall i \in \{1,...,n\},
\end{equation}
where $n = |X|$. Consequently, easier images have a higher chance of being sampled in the first learning iterations. When $k > 1$ in Eq.~\eqref{eq_scoring_function}, we need to add the constant $k-1$ to each term in Eq.~\eqref{eq_prob}, i.e. we replace $w_P(x,t)$ with $w_P(x,t) + k - 1$, to obtain positive values. Towards the end of the training process, as $w_P$ converges asymptotically to $1$ (see Eq.~\eqref{eq_scoring_convergence}), it becomes equally likely to sample an easy or a difficult image, i.e.:
\begin{equation}
\lim_{w_P \rightarrow{1}} Prob(R=i) = \frac{1}{n}, \forall i \in \{1,\dots,n\},
\end{equation}
where $n = |X|$. At the limit, $p_{w_P}$ converges to a uniform density and $p_{r,w_P}$ becomes equal to $p_r$.

\noindent 
{\bf Observation.} $V^{(2)}$ can be seen as a continuous version of our first curriculum learning approach, in which the probability of sampling a real image $x$ from the set of training images $X$ is given by a step function, where the number of steps is equal to the number of batches $m$.

\vspace{-0.1cm}
\section{Experiments}
\label{sec_Experiments}
\vspace{-0.1cm}
\subsection{Data sets}
\vspace{-0.1cm}

We perform image generation experiments on the CIFAR-10 data set~\cite{Krizhevsky-TR-2009}. It consists of 50000 color train images of $32 \times 32$ pixels, equally distributed into 10 classes: airplane, automobile, bird, cat, deer, dog, frog, horse, ship and truck. 
Our translation experiments include two of the data sets used in~\cite{Zhu-ICCV-2017}. Horse2zebra contains 939 horse images and 1177 zebra images downloaded from ImageNet~\cite{Russakovsky2015} using the keywords \emph{wild horse} and \emph{zebra}. Apple2orange has 996 apple images and 1020 orange images from the same source, labeled with \emph{apple} and \emph{navel orange}. All images are $256 \times 256$ pixels in size.

\vspace{-0.1cm}
\subsection{Baselines, evaluation and parameter choices}
\vspace{-0.1cm}

\noindent 
{\bf Baselines.}
For the image generation experiments on CIFAR-10, we employ a state-of-the-art baseline, SNGAN~\cite{Miyato-ICLR-2018}, which is based on the Hinge loss. We consider SNGAN as the most relevant baseline, since we use it as starting point for our curriculum learning approaches. However, we include additional models from the recent literature, namely DCGAN~\cite{Radford-ICLR-2016}, WGAN-GP~\cite{Gulrajani-NIPS-2017}, Parallel Optimal Transport GAN (POT-GAN)~\cite{Avraham-CVPR-2019} and Generative Latent Nearest Neighbors (GLANN)~\cite{Hoshen-CVPR-2019}. We also include the results of the acGAN proposed by Doan et al.~\cite{Doan-AAAI-2019}, which uses adaptive curriculum. Since our curriculum SNGAN-based models are unsupervised, we do not compare with class conditional (supervised) baselines.
For the image translation experiments, we employ CycleGAN~\cite{Zhu-ICCV-2017} as baseline. 

\noindent 
{\bf Evaluation metrics.}
Evaluating the quality and realism of generated content as perceived by humans is not an easy task. At this moment, there is no universally agreed metric able to measure the outputs of GANs, each having its own shortcomings. To automatically quantify the performance, we use the Inception Score (IS)~\cite{Salimans-NIPS-2016} and the Fr\'echet Inception Distance (FID)~\cite{Heusel-NIPS-2017}, which are computed over $10000$ generated images (not used in the training process). The reported scores are averaged over $5$ runs. A higher IS or a lower FID indicates higher performance.
Along with the automatic metrics, we evaluate the results by asking humans to annotate images, in order to determine if they prefer the baseline GANs or our Curriculum-GANs (CuGANs). 


\noindent 
{\bf Implementation details.}
In the image generation experiments, we used the SNGAN implementation available at \url{https://github.com/watsonyanghx/GAN\_Lib\_Tensorflow}, which can reproduce the results reported in~\cite{Miyato-ICLR-2018}. The model is based on ResNet. We trained the model for 80000 iterations using mini-batches of 64 samples. We observed that the Inception Score stabilizes much sooner (before 50000 iterations) using the Adam optimizer~\cite{Kingma-ICLR-2015}. The learning rate is $2 \cdot 10^{-4}$. For the first curriculum learning approach, we split the training set in $m=3$ batches, as Ionescu et al.~\cite{Ionescu-CVPR-2016}: an easy batch, a medium batch, and a difficult batch. Each batch contains the same number of samples. For the second and the third curriculum learning approaches, we set $\gamma = 5 \cdot 10^{-5}$, which is chosen with respect to the total number of iterations (80000). We conducted preliminary experiments to tune the other parameters. For the first curriculum learning approach, we experimented with three different numbers of iterations to train on the easy batch (5000, 10000 and 15000), and another three numbers of iterations to train on the easy and medium batches together (15000, 20000 and 25000). We obtained slightly better results for training on the easy batch for 15000 iterations, and on both easy and medium batches for 25000 iterations. The rest of the iterations (40000) include all three batches in the training. For the second and the third curriculum learning approaches, we conducted experiments with $k \in \{1, 2, 4\}$. When we weight the training images with the corresponding difficulty scores (as in Eq.~\eqref{eq_curriculum2}), we obtain optimal results with $k=2$. When we sample the training images according to the difficulty scores (as in Eq.~\eqref{eq_curriculum3}), we obtain optimal results with $k=4$.

In the image translation experiments, we used the CycleGAN implementation available at \url{https://github.com/leehomyc/cyclegan-1}. The model is trained for 25000 iterations, using a mini-batch size of $8$ samples. As for SNGAN, we employ the Adam optimizer~\cite{Kingma-ICLR-2015} with a learning rate of $2 \cdot 10^{-4}$. The weight of the cycle consistency loss term in the full objective function is set to $\lambda = 10$, as in the original paper~\cite{Zhu-ICCV-2017}. We apply linear weight decay after the first 12500 iterations. We compare the baseline CycleGAN with the Curriculum-CycleGAN based on weighting the training images with the corresponding difficulty scores, since the weighting strategy provides the best FID score in the image generation experiments on CIFAR-10. We did not evaluate the other two curriculum learning approaches to avoiding tripling the human annotation time and costs.

\vspace{-0.1cm}
\subsection{Image generation results}
\vspace{-0.1cm}

\begin{figure}[!t]
\begin{center}
\includegraphics[width=1.0\linewidth]{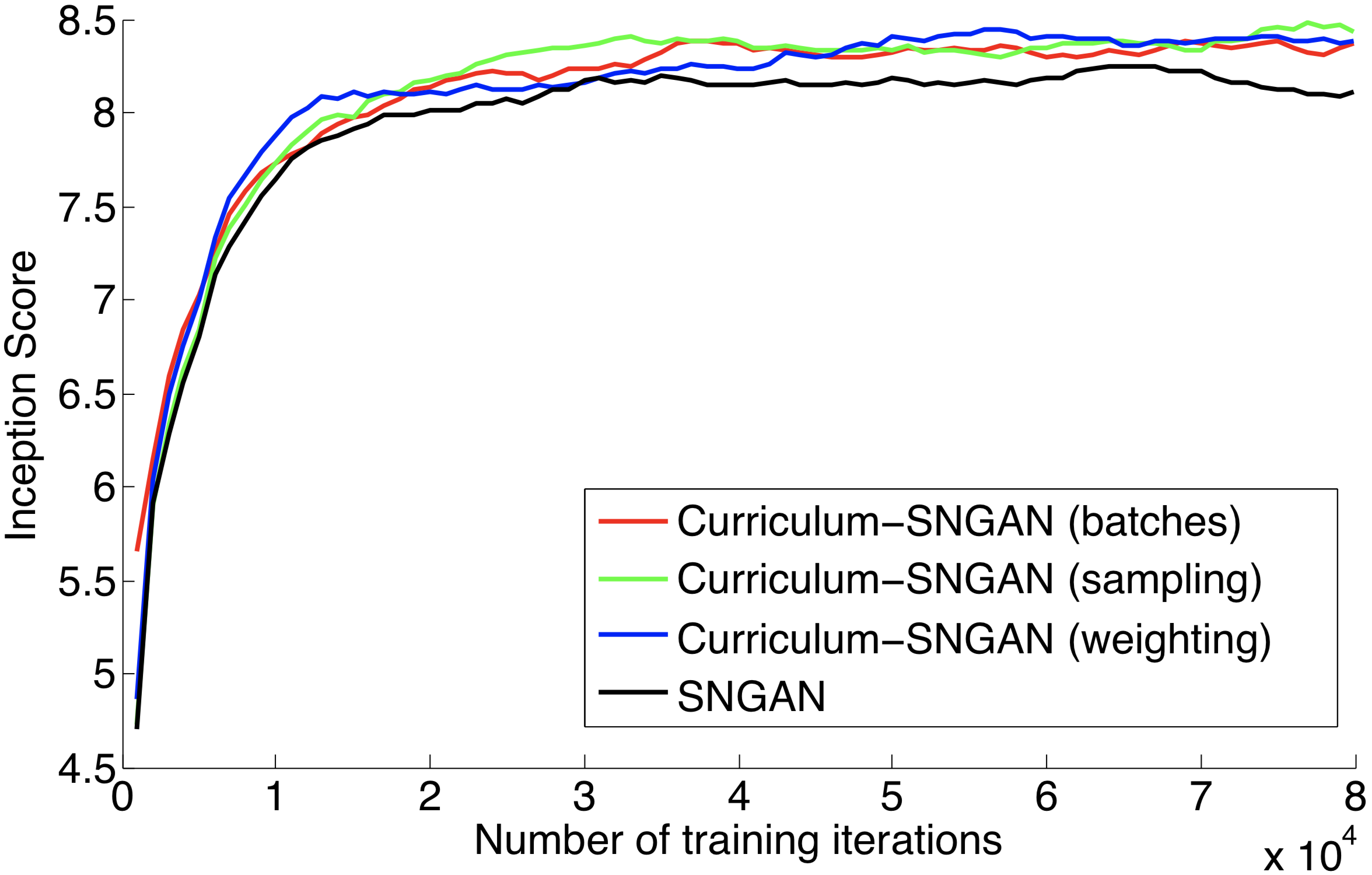}
\end{center}
\vspace*{-0.3cm}
\caption{Inception Scores (IS) of SNGAN (baseline) versus three Curriculum-SNGAN models based on various curriculum learning strategies (batches, sampling, weighting), on CIFAR-10. The scores are computed on generated images, not used in the training process. Best viewed in color.}
\label{fig_CIFAR_IS}
\end{figure}


\begin{table}[t]
\setlength\tabcolsep{2.8pt}
\small{
\begin{center}
\begin{tabular}{|l|c|c|}
\hline
Method                                              & IS                & FID   \\
\hline
\hline
DCGAN~\cite{Radford-ICLR-2016}                      & $6.16\pm0.07$     & $71.07\pm1.06$ \\
DCGAN (with ResNet)$^*$~\cite{Radford-ICLR-2016}    & $6.64\pm0.14$     & - \\
WGAN-GP$^*$~\cite{Miyato-ICLR-2018}                 & $6.68\pm0.06$     & $40.20$ \\
WGAN-GP (with ResNet)~\cite{Gulrajani-NIPS-2017}    & $7.86\pm0.07$     & - \\
GLANN~\cite{Hoshen-CVPR-2019}                       & -                 & $46.50\pm 0.20$ \\
POT-GAN~\cite{Avraham-CVPR-2019}                    & $6.87\pm0.04$     & $32.50$ \\
acGAN~\cite{Doan-AAAI-2019}                         & $6.22\pm0.04$     & $49.81\pm0.23$ \\
\hline
SNGAN$^*$~\cite{Miyato-ICLR-2018}                   & $8.22\pm0.05$     & $21.70\pm0.21$ \\
\hline
Curriculum-SNGAN (batches)                          & $8.46\pm0.13$     & $14.64\pm0.31$ \\
Curriculum-SNGAN (weighting)                        & $8.44\pm0.11$     & $\mathbf{14.41}\pm0.24$ \\
Curriculum-SNGAN (sampling)                         & $\mathbf{8.51}\pm0.09$ & $14.48\pm0.26$ \\
\hline
\end{tabular}
\end{center}
\vspace*{-0.15cm}
\caption{Inception Scores (IS) and Fr\'echet Inception Distances (FID) on CIFAR-10. Several unsupervised GAN models~\cite{Avraham-CVPR-2019,Doan-AAAI-2019,Gulrajani-NIPS-2017,Hoshen-CVPR-2019,Miyato-ICLR-2018,Radford-ICLR-2016} are compared with our SNGAN-based approaches, each employing a different curriculum learning strategy proposed in this paper. The results marked with an asterisk are taken from the SNGAN paper~\cite{Miyato-ICLR-2018}. The results of acGAN are based on the source code provided by Doan et al.~\cite{Doan-AAAI-2019}. The best IS (higher is better) and the best FID (lower is better) scores are marked in bold.\label{tab_results_CIFAR}}
}
\vspace*{-0.1cm}
\end{table}

\begin{figure*}[!t]
\begin{center}
\includegraphics[width=0.94\linewidth]{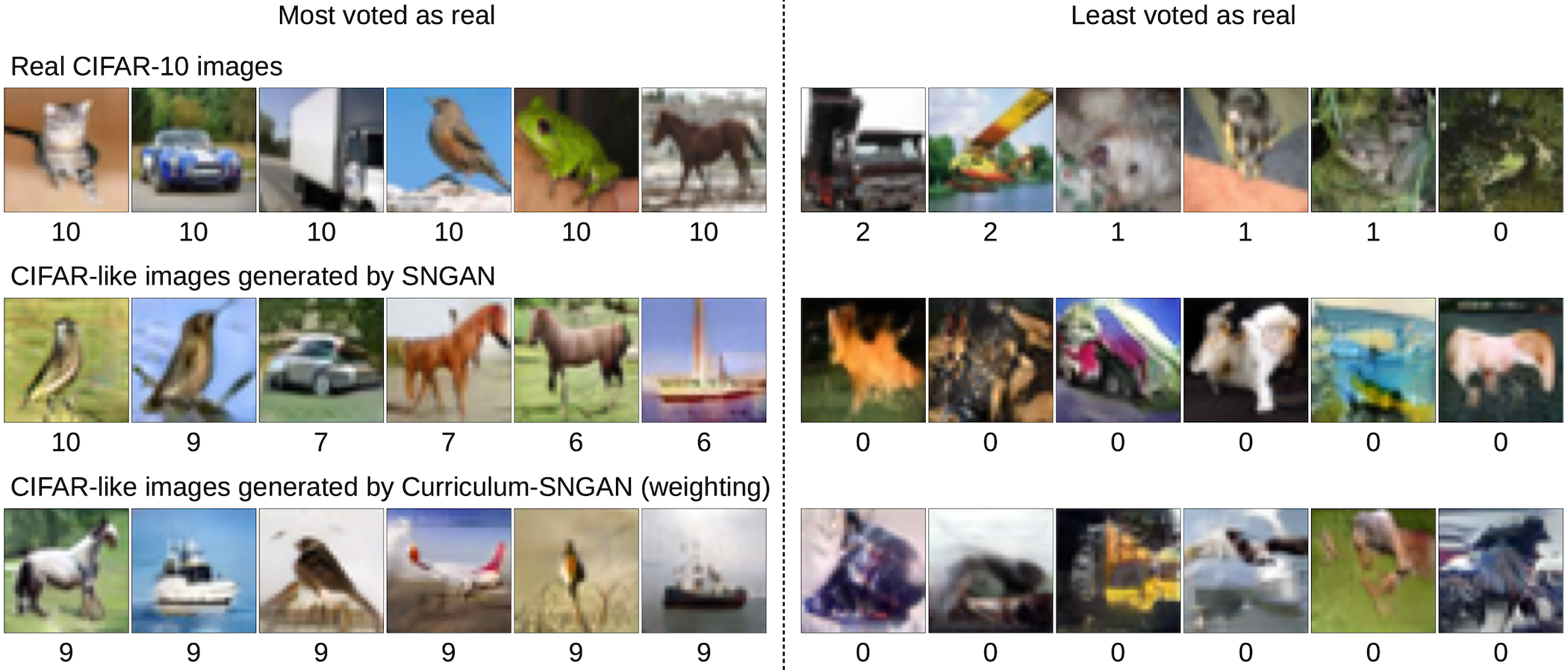}
\end{center}
\vspace*{-0.4cm}
\caption{Most voted and least voted images from the set of 600 images labeled by human annotators. Images on each row are selected from different subsets: real images, generated by SNGAN and generated by Curriculum-SNGAN with weighting. Best viewed in color.}
\label{fig_CIFAR_voted}
\vspace*{-0.4cm}
\end{figure*}

\noindent
{\bf Faster convergence.}
In Figure~\ref{fig_CIFAR_IS}, we present the evolution of the Inception Scores for the standard SNGAN and three other SNGAN versions, that are enhanced through one of our curriculum learning strategies, on CIFAR-10. We note that the performance of each model stabilizes after 50000 iterations. However, the curriculum-based models reach higher IS values, right from the first iterations. This indicates that the Curriculum-SNGANs converge faster than the standard SNGAN. For instance, the Curriculum-SNGAN based on sampling (corresponding to Eq.~\eqref{eq_curriculum3}) achieves about the same IS value as the baseline SNGAN, in only 20000 iterations instead of 65000 iterations.

\begin{table}[t]
\setlength\tabcolsep{1.2pt}
\small{
\begin{center}
\begin{tabular}{|l|c|c|c|c|c|c|c|c|c|c|c|}
\hline
                & A & B & C & D & E & F & G & H & I & J & Avg. \\
\hline
\hline
CIFAR-10        & $156$ & $195$ & $124$ & $168$ & $151$ & $142$ & $167$ & $160$ & $95$  & $127$ & $74.3\%$ \\
\hline
SNGAN~\cite{Miyato-ICLR-2018}   & $36$  & $111$ & $26$  & $18$  & $30$  & $34$  & $40$  & $38$  & $24$  & $11$  & $18.4\%$ \\
\hline
Curriculum      & $48$  & $123$ & $27$  & $27$  & $39$  & $60$  & $67$  & $58$  & $28$  & $23$  & $25.0\%$ \\
\hline
\end{tabular}
\end{center}
\vspace*{-0.15cm}
\caption{Number of images voted as real by 10 human annotators (identified by letters from A to J). The annotators were asked to label 600 images (200 real CIFAR-10 images, 200 images generated by SNGAN and another 200 images generated by Curriculum-SNGAN with weighting) as real or fake.\label{tab_results_human_CIFAR}}
}
\vspace*{-0.4cm}
\end{table}

\noindent
{\bf Superior results.}
In Table~\ref{tab_results_CIFAR}, we compare our Curriculum-SNGANs with the standard SNGAN, as well as DCGAN~\cite{Radford-ICLR-2016}, WGAN-GP~\cite{Gulrajani-NIPS-2017}, GLANN~\cite{Hoshen-CVPR-2019}, POT-GAN~\cite{Avraham-CVPR-2019} and acGAN~\cite{Doan-AAAI-2019}, on CIFAR-10. First, we note that SNGAN achieves the best results among all baselines, confirming that SNGAN is indeed representative for the state-of-the-art. We observe that all our curriculum learning strategies can further boost the performance of SNGAN. When we divide the images into easy-to-hard batches, we achieve an IS of $8.46$. The best IS score ($8.51$) is obtained when we sample the train images according to difficulty. For an unsupervised model, we believe that an IS of $8.51$ is noteworthy. Furthermore, our improvements in terms of FID are much higher than all baselines, even compared to the adaptive curriculum approach of Doan et al.~\cite{Doan-AAAI-2019}. While easy-to-hard batches and sampling provide better IS values, we observe that the Curriculum-SNGAN based on weighting according to difficulty (corresponding to Eq.~\eqref{eq_curriculum2}) achieves the best FID value ($14.41$). For this reason we choose this curriculum learning strategy for the human evaluation experiments.

We asked 10 human annotators to label images either as real or fake. We provided the same set of 600 images (presented in a random order) to each annotator. We randomly selected 200 real CIFAR images, 200 images generated by SNGAN and 200 images generated by Curriculum-SNGAN (weighting). The goal of the annotation study is to determine the percentage of generated images that fool the annotators, using the real images as a control set (preventing evaluators from labeling every image as fake). In Table~\ref{tab_results_human_CIFAR}, we report the number of images labeled as real by each annotator. We note that in $25.7\%$ cases, the annotators labeled real CIFAR images as being fake. Nevertheless, the humans largely figured out what images are generated. The standard SNGAN fooled annotators in $18.4\%$ cases, while the Curriculum-SNGAN fooled annotators in $25.0\%$ cases. Interestingly, each and every human labeled more images generated by Curriculum-SNGAN as real than images generated by the baseline SNGAN. As illustrated in Figure~\ref{fig_CIFAR_voted}, there are several images generated by Curriculum-SNGAN, which are labeled as real by 9 out of 10 annotators. The number of votes drops faster for the standard SNGAN approach. All results indicate that the images generated by Curriculum-SNGAN are superior to those generated by SNGAN.

\vspace{-0.1cm}
\subsection{Image translation results}
\vspace{-0.1cm}

\begin{table}[t]
\setlength\tabcolsep{2.0pt}
\small{
\begin{center}
\begin{tabular}{|l|c|c|c|c|c|}
\hline
Option & H$\rightarrow$Z & Z$\rightarrow$H & A$\rightarrow$O & O$\rightarrow$A  & Avg. \\
\hline
\hline
CycleGAN~\cite{Zhu-ICCV-2017}  & $11.9\%$ & $13.9\%$ & $20.6\%$ & $32.7\%$ & $19.8\%$\\
\hline
Curriculum-CycleGAN     & $52.5\%$ & $37.4\%$ & $37.1\%$ & $35.1\%$ & $40.5\%$\\
\hline
Ties                    & $35.6\%$ & $48.7\%$ & $42.3\%$ & $32.2\%$ & $39.7\%$\\
\hline
\end{tabular}
\end{center}
\vspace*{-0.15cm}
\caption{Average percentage of cases in which 6 human annotators consider images generated by CycleGAN as better, images generated by Curriculum-CycleGAN (weighting) as better, or both equally good. Evaluations are provided for 4 test sets of images: horse2zebra (H$\rightarrow$Z), zebra2horse (Z$\rightarrow$H), apple2orange (A$\rightarrow$O), orange2apple (O$\rightarrow$A). The overall average is also included.\label{tab_results}}
}
\vspace*{-0.4cm}
\end{table}

\begin{figure*}[!t]
\begin{center}
\includegraphics[width=0.94\linewidth]{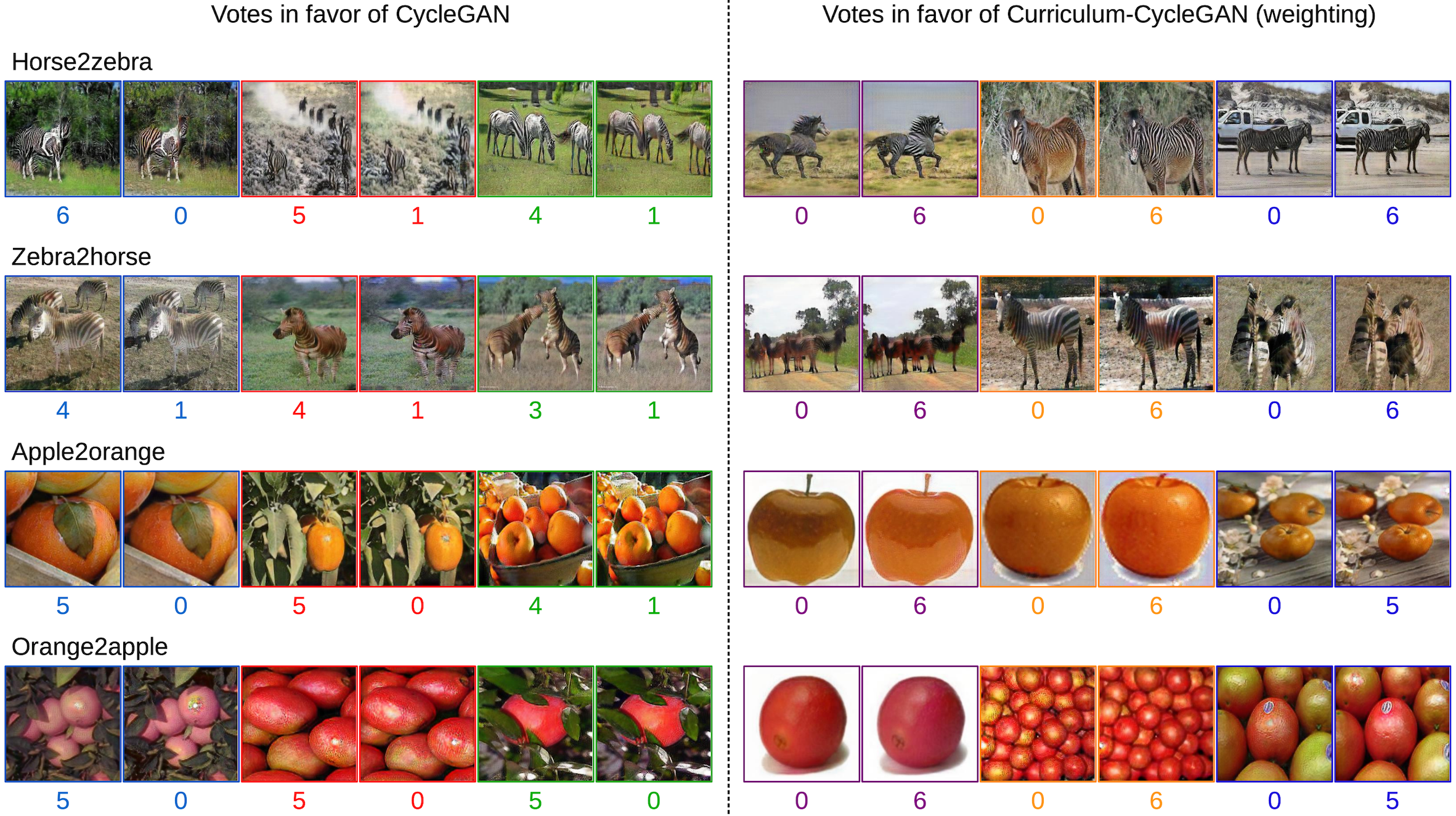}
\end{center}
\vspace*{-0.45cm}
\caption{Side by side image pairs generated by CycleGAN (left image in each pair) and Curriculum-CycleGAN (right image in each pair) with the corresponding number of votes provided by 6 human annotators. When the sum of the number of votes in a pair is lower than 6, it means that the missing votes correspond to ties. Image pairs that received most votes in favor of CycleGAN are presented in the left-hand side of the figure, while image pairs that received most votes in favor of Curriculum-CycleGAN are presented in the right-hand side. Best viewed in color.}
\label{fig_translation_voted}
\vspace*{-0.3cm}
\end{figure*}

The image translation results are evaluated only by human annotators. There are four test sets of images~\cite{Zhu-ICCV-2017}, corresponding to the following translations: horse2zebra (120 images), zebra2horse (140 images), apple2orange (266 images) and orange2apple (248 images). We asked 6 human annotators to choose between the images translated by CycleGAN and those translated by Curriculum-CycleGAN (weighting), without disclosing any information about the models. In each case, we also provided the original (source) image. Since the test images are fixed for both models, the random chance factor is eliminated. In Figure~\ref{fig_translation_voted}, we show several images translated by both models side by side. We notice that there are several image pairs in which all 6 annotators opted for Curriculum-CycleGAN. For horse2zebra, the baseline CycleGAN wins when our model produces brownish zebras. For apple2orange, annotators prefer the baseline when our model produces artifacts, but they prefer our model when it produces the right tone of orange. 

In Table~\ref{tab_results}, we present the average percentage of cases (computed on the 6 annotators) in which the annotators prefer either the CycleGAN output images or the Curriculum-CycleGAN output images, as well as the percentage of tied cases (images are labeled as equally good). We note that on three sets of images (horse2zebra, zebra2horse and apple2orange), the annotators show significant preference for our Curriculum-CycleGAN based on weighting. Furthermore, in these three test sets, all humans prefer our model over the baseline CycleGAN (the individual percentages are not shown in Table~\ref{tab_results} due to lack of space). For orange2apple, only 2 out of 6 annotators prefer our model, although our model has a higher average preference ($35.1\%$) compared to the baseline ($32.7\%$), as seen in Table~\ref{tab_results}. All in all, the human annotators seem to prefer the curriculum-based approach in $20.7\%$ more cases than the baseline CycleGAN, confirming once more that the curriculum strategy is indeed useful. 

\vspace{-0.2cm}
\section{Conclusion}
\label{sec_Conclusion}
\vspace{-0.1cm}

In this paper, we presented three curriculum learning strategies for training GANs. The empirical results indicate that our curriculum learning strategies achieve faster convergence during training, i.e. the number of training iterations can be reduced by a factor of three without affecting the quality of the generative results. Furthermore, using a similar number of training iterations, our curriculum learning strategies can boost the quality of the generative and translation results, surpassing all considered baselines~\cite{Avraham-CVPR-2019,Doan-AAAI-2019,Gulrajani-NIPS-2017,Hoshen-CVPR-2019,Miyato-ICLR-2018,Radford-ICLR-2016,Zhu-ICCV-2017} on CIFAR-10. Both automatic measures and human evaluators confirm our findings. While we conducted experiments on images of $32 \times 32$ and $256 \times 256$ of pixels in size, in future work, we aim to apply our curriculum learning strategies in order to generate larger images, containing a natural variety of object classes.

\noindent
{\bf Acknowledgements.}
Work funded from a grant of Ministry of Research and Innovation, CNCS-UEFISCDI, project no. PN-III-P1-1.1-PD-2016-0787, and from the EEA Grants 2014-2021, project no. EEA-RO-NO-2018-0496.

{\small
\bibliographystyle{ieee}
\bibliography{references}
}

\end{document}